\documentclass[letterpaper, 10 pt, conference]{ieeeconf}  

\IEEEoverridecommandlockouts                              

\title{\LARGE \bf
Building Transportation Foundation Model via Generative Graph Transformer
}

\author{Xuhong Wang$^{1,\dagger}$, Ding Wang$^{1,\dagger}$, Liang Chen$^{1}$ and Yilun Lin$^{1,*}$, \textit{Member, IEEE}
\thanks{This work is supported by the Shanghai Artificial Intelligence Laboratory.}
\thanks{$\dagger$ Equal contribution.}
\thanks{*Corresponding author: Yilun Lin (linyilun@pjlab.org.cn)}
\thanks{$^{1}$Xuhong Wang (wangxuhong@pjlab.org.cn), Ding Wang (wangding@pjlab.org.cn), Liang Chen (chenliang@pjlab.org.cn), Yilun Lin (linyilun@pjlab.org.cn) are with Urban Computing Lab, Shanghai Artificial Intelligence Laboratory, Shanghai, China.}%
}

\usepackage{graphicx}
\usepackage{enumerate}
\usepackage{amsmath}
\usepackage{url}
\usepackage{hyperref}

\newtheorem{definition}{Definition}[section]

\begin{document}

\maketitle
\thispagestyle{empty}
\pagestyle{empty}

\begin{abstract}

Efficient traffic management is crucial for maintaining urban mobility, especially in densely populated areas where congestion, accidents, and delays can lead to frustrating and expensive commutes. However, existing prediction methods face challenges in terms of optimizing a single objective and understanding the complex composition of the transportation system. Moreover, they lack the ability to understand the macroscopic system and cannot efficiently utilize big data. In this paper, we propose a novel approach, Transportation Foundation Model (TFM), which integrates the principles of traffic simulation into traffic prediction. TFM uses graph structures and dynamic graph generation algorithms to capture the participatory behavior and interaction of transportation system actors. This data-driven and model-free simulation method addresses the challenges faced by traditional systems in terms of structural complexity and model accuracy and provides a foundation for solving complex transportation problems with real data. The proposed approach shows promising results in accurately predicting traffic outcomes in an urban transportation setting.

\end{abstract}

\section{INTRODUCTION}

The most promising approach to creating intelligent transportation systems is the utilization of spatio-temporal data to predict traffic patterns. Therefore, the issue of traffic prediction~\cite{liu2018traffic} has gained significant attention from both academic and industry professionals. Traffic prediction plays a crucial role in alleviating traffic congestion, enhancing traffic safety, optimizing urban planning, and improving citizens' travel experiences \cite{jiang2022cellular}. 

However, there are two problems with current traffic prediction technology. Firstly, this technology mainly targets the prediction of certain fixed indicators, including origin-destination travel time \cite{li2020hierarchical}, path-travel time \cite{xing2013designing}, travel demand \cite{yuan2021effective}, regional flow \cite{nagy2018survey}, network flow \cite{lohrasbinasab2022statistical}, and traffic speed \cite{asif2013spatiotemporal}, which cannot establish a common knowledge system for intelligent transportation systems. A well-known example is the large language model (LLM) \cite{beltagy2019scibert}, such as ChatGPT \cite{lund2023chatting}, whose training objective never includes a specific metric, but instead uses generative machine learning techniques to estimate the probability distribution of all real-world language data. 

The first major issue is the training objective, while the second concerns the granularity problem of the traffic prediction model. As we all know, the spatial-temporal data used in traffic prediction models usually come from cameras, magnetic sensors, and smartphones in the road network. This data has a coarse granularity and cannot model micro-interaction mechanisms between vehicles. However, the macroscopic traffic system is composed of many microscopic subsystems. Studying vehicle interactions at a microscopic scale can help people gain a deeper understanding of the overall traffic system.

We believe that the most essential solution to the above two problems is to integrate some concepts from traffic simulation into traffic prediction, bridging the gap between these two research fields. The main feature of traffic prediction technology is specificity. It is usually used to predict the spatio-temporal evolution of a limited number of traffic indicators, while the concept of traffic simulation technology has a fundamental difference from traffic prediction. 

Traffic simulation uses mathematical models to describe the behavior of individuals or the entire system to reflect complex traffic phenomena, helping traffic managers better understand traffic evolution. This is a bottom-up modeling approach that not only focuses on a single prediction target, but also leans towards the modeling of the entire transportation system. Traffic simulation involves the development of rules and functions based on expert knowledge and assumptions to simulate the traffic network. Traditionally, these rules and functions are based on expert knowledge and assumptions about traffic behavior, as well as limited real-world data. This has led to traffic simulation being unable to leverage the large-scale data available today to break through the bottleneck of modeling accuracy.

\begin{figure}[htp]
\centering
\includegraphics[width=1\linewidth]{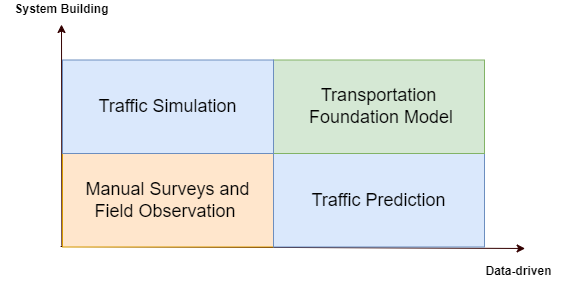}
\caption{Different methods used for modeling transportation systems. The most primitive method of modeling transportation systems was manual surveys and field observation. Traffic simulation tends to model the behavior of each traffic participant through physical mechanisms to ultimately form an understanding of the entire traffic system. Traffic prediction typically uses a fully data-driven approach to predict indicators in certain perspectives or scenarios within the transportation system, without generating a comprehensive understanding of the entire system.}
\label{fig:Intro}
\end{figure}

Following the large models used in natural language processing, we attempt to establish a basic model in the field of transportation. This model is not intended to predict a specific metric in traffic, but rather to explore how to model traffic interactions from the perspectives of various traffic participants and ultimately establish a large model that can understand the entire traffic system, and improve the performance of various traffic prediction and simulation tasks.

We propose Transportation Foundation Model (TFM) to capture the participants and their interactions in the transportation system using the graph structure and dynamic graph generation algorithm. Each participant in the transportation system (such as vehicles, pedestrians, smart traffic lights, etc.) is represented as a node, and their relationships are represented as edges, which together form a traffic graph representing the state of all participants in the current transportation system. The temporal changes of the traffic graph can represent the evolution of the entire traffic system, and this data organization method provides a more adaptive and flexible way to model complex transportation problems. This data-driven, and model-free simulation system approach can address the challenges of structural complexity and model accuracy in traditional systems, and provide a basis for complex traffic problems with real data.

This study provides an overview of proposed Transportation Foundation Model. Section 2 introduces related work in traffic simulation. Section 3 provides basic introductions to the concepts and models used in TFM. Section 4 presents a general introduction to the structure and workflow of TFM. Section 5 provides an example scenario to demonstrate the capabilities of TFM. Finally, the study concludes with an outlook on future work, highlighting possible directions for further research and development.

\section{RELATED WORKS}
\subsection{Traffic Prediction}
Traffic flow forecasting is crucial in Intelligent Transportation Systems (ITS). Two main methods are dynamic modeling and data mining. Dynamic modeling uses complicated formulations and theoretical assumptions to predict traffic dynamics, while data mining finds historical data patterns for predicting future trends. Early data mining models like Historical Average (HA) are simple and fast but have low accuracy. Newer models, like ARIMA~\cite{saadallah2018bright}, VAR~\cite{chen2022sparse}, and Kalman filter~\cite{kumar2017traffic}, require low computational resources but assume stationarity and can't deal with non-linearity. Nonparametric and Learnable models like support vector regression~\cite{ahn2016highway}, k-nearest neighbor~\cite{cai2016spatiotemporal}, and Bayesian network~\cite{sun2015dynamic} have emerged to tackle these problems. 

With the development of deep learning in various fields, researchers have applied deep learning models in transportation systems~\cite{yao2019revisiting}. Deep belief networks~\cite{nie2017network} extract high-dimensional features but can't extract specific spatiotemporal features. An RNN~\cite{qiu2018spatio} can better learn temporal dependence, but causes gradient problems. LSTM and GRU~\cite{fu2016using} are improvements but have limitations in performance when stacked. A CNN~\cite{yang2019mf} has stable gradients and low memory requirements, but requires a stack of multiple layers and doesn't naturally consider spatial dependence. ST-ResNet~\cite{zhang2018predicting} focuses on temporal closeness, period, and trend but CNN format needs to be changed, and the captured spatial features remain incomplete. Dilated convolution networks, like WaveNet~\cite{tian2021spatial}, can extract longer sequence information without increasing parameters and are suitable for medium- and long-term traffic prediction.

\subsection{Graph Learning in Traffic Prediction}
The study of traffic forecasting is an emerging topic in graph learning with significant potential impact on daily life.  DCRNN~\cite{lidiffusion} combines graph convolution with recurrent neural networks in an encoder-decoder manner, while STGCN~\cite{yu2018spatio} combines graph convolution with gated temporal convolution. Graph WaveNet~\cite{wu2019graph} combines adaptive graph convolution with dilated casual convolution to capture spatiotemporal dependencies, while ASTGCN~\cite{guo2019attention} uses both the spatial-temporal attention mechanism and the spatial-temporal convolution. STG2Seq~\cite{bai2019stg2seq} uses multiple gated graph convolutional modules and seq2seq architecture with attention mechanisms to make multi-step predictions. STSGCN~\cite{song2020spatial} utilizes multiple localized spatial-temporal subgraph modules to capture localized spatial-temporal correlations directly. LSGCN~\cite{huang2020lsgcn} integrates a novel attention mechanism and graph convolution into a spatially gated block. AGCRN~\cite{bai2020adaptive} utilizes node adaptive parameter learning to capture node-specific spatial and temporal correlations in time-series data automatically without a pre-defined graph, whereas STFGNN ~\cite{li2021spatial} captures hidden spatial dependencies through a novel fusion operation of various spatial and temporal graphs that are treated for different time periods in parallel. Finally, Z-GCNETs \cite{chen2021z} integrate a new time-aware zigzag topological layer into time-conditioned GCNs, and STGODE \cite{fang2021spatial} captures spatiotemporal dynamics through a tensor-based ODE.

\subsection{Traffic Simulation}
A traffic simulation tool for urban road networks is presented in 1997, which is based on the Nagel–Schreckenberg Model \cite{esser1997microscopic}. Genetic algorithms have been applied to airport ground traffic optimization \cite{gotteland2003genetic}. A ground traffic simulation tool is proposed and applied to Roissy Charles De Gaulle airport. Truly agent-based traffic and mobility simulations have been studied \cite{balmer2004towards}. A real-world case study is presented, which says that multi-agent methods for traffic are mature enough to be used alongside existing methods. ParamGrid, a scalable and synchronized framework, was presented, and it can distribute the simulation across a cluster of ordinary-performance, low-cost personal computers (PCs) connected by local area network (LAN) \cite{klefstad2005distributed}. A microscopic model was proposed, which is able to simulate traffic situations in an urban environment in real time for use in driving simulators \cite{maroto2006real}. The functionality of simulation models is enhanced, and reliability is improved The proposed approach offers significant advantages over conventional methods where historical and outdated data is used to run poorly calibrated traffic simulation models \cite{puri2007statistical}. A multi-agent behavioral model was proposed, and it is based on (i) the opportunistic individual behaviors that describe the norm violation and (ii) the anticipatory individual abilities of simulated drivers that allow critical situations to be detected \cite{doniec2008behavioral}. A novel, real-time algorithm for modeling large-scale has been presented, realistic traffic using a hybrid model of both continuum and agent-based methods for traffic simulation \cite{sewall2011interactive}. The latest developments concerning intermodal traffic solutions, simulator coupling and model development and validation were presented on the example of the open-source traffic simulator SUMO \cite{lopez2018microscopic}.

\section{METHODS}
\subsection{From Traffic Networks to Transportation Graphs}

The concept of the Dynamic Transportation graph defined in this article is not the same as the traffic network mentioned in other literature. This is because the traffic network usually only includes elements such as lanes and intersections to form a road network, while Dynamic Transportation Graph not only includes the traffic network, but also describes each participant in the traffic system, such as vehicles, pedestrians, traffic lights, etc.

\begin{definition}[Dynamic Transportation Graph]
A traffic system can be described in a Dynamic Graph (DG)$\mathcal{G}_{n}(\mathcal{V}_{n}, \mathcal{E}_{n}, \mathbf{S}_n, \mathbf{A}_n)$, where $n = 1,2,3,\dots$ represents the evolving steps of the system and $\mathcal{V}_{n}$ represents all participants in it. $\mathcal{E}_{n} \subseteq \mathcal{V}_{n} \times \mathcal{V}_{n}$ is the relationship between those traffic participants$e_i =(u_i,v_i ) \in \mathcal{E}_{n}   ,u_i   ,v_i    \in \mathcal{V}_{n}$, which means each relational edge consists of a source node $u_i$ and a target node$v_i$. The nodes and edges of a dynamic graph may have feature attributes attached, $\mathbf{S}_n\in \mathcal{R}^{|\mathcal{V}_n| \times D}$ represents the node state matrix of$\mathcal{V}_{n}$, where$D$ is the dimension of the node state vector. The state attribute of each single node $v_i \in \mathcal{V}_n$ is defined as $\mathbf{s}_{i} \in \mathcal{R}^{D}$. Generally speaking, the features on the node represent the current state of traffic participants, such as current vehicle speed, acceleration, traffic light status, traffic flow of the road, etc., and$\mathbf{A}_n\in \mathcal{R}^{|\mathcal{V}_n| \times |\mathcal{V}_n|}$ represents the adjacency matrix of $\mathcal{V}_{n}$. 
\end{definition}

\begin{definition}[Dynamic Graph Transformation]
According to the definition of Dynamic Transportation Graph, the overall dynamic evolution process of traffic system can be regarded as a dynamic graph transformation process $\mathcal{T}: \mathcal{G}_{n} \rightarrow \mathcal{G}_{n+1}$. Essentially, this process is to learn a conditional probability function, which can be modeled as a probability chain.
\begin{equation}
p\left(\mathcal{G}_{n+1}\right) = \prod^N_{n=1} p\left(\mathcal{G}_{n+1} \mid \mathcal{G}_{n}\right) 
\end{equation}
\end{definition}
The parameters of the model are learned from the training data using the maximum likelihood estimation method, which maximizes the likelihood function of the training data.

\subsection{The Framework of Transportation Foundation Model}
Due to the difficulty of directly modeling the process of Dynamic Graph Transformation, it can be decomposed into three related sub-tasks:

\begin{enumerate}
    \item \textbf{Dynamic graph representation} $p\left( \mathbf{Z}_n \mid \mathcal{G}_{n}\right)$: As all data can be represented as a graph, learning from historical data is essentially a feature learning of dynamic graphs. Its goal is to encode the knowledge and information contained in historical dynamic graphs into graph feature vectors $\mathbf{Z}_{n}$.
   \item \textbf{Interaction generation} $ p\left(u_i, v_i \mid \mathbf{Z}_n \right)$: The interaction behavior between any two traffic participants can be represented as an edge on a dynamic graph. The generation of the interaction is to obtain the joint probability distribution of the edges $e_i   =(u_i  ,v_i   )$.
   \item \textbf{State prediction} $ p\left(\mathbf{s}_{i} \mid \mathbf{Z}_n \right) $: The state change of traffic participants can be represented as a feature change of nodes on the graph. The model needs to use historical status and relational information to predict the state information of each participant in the next step.
\end{enumerate}

\begin{figure*}[htp]
\centering
\includegraphics[width=0.8\linewidth]{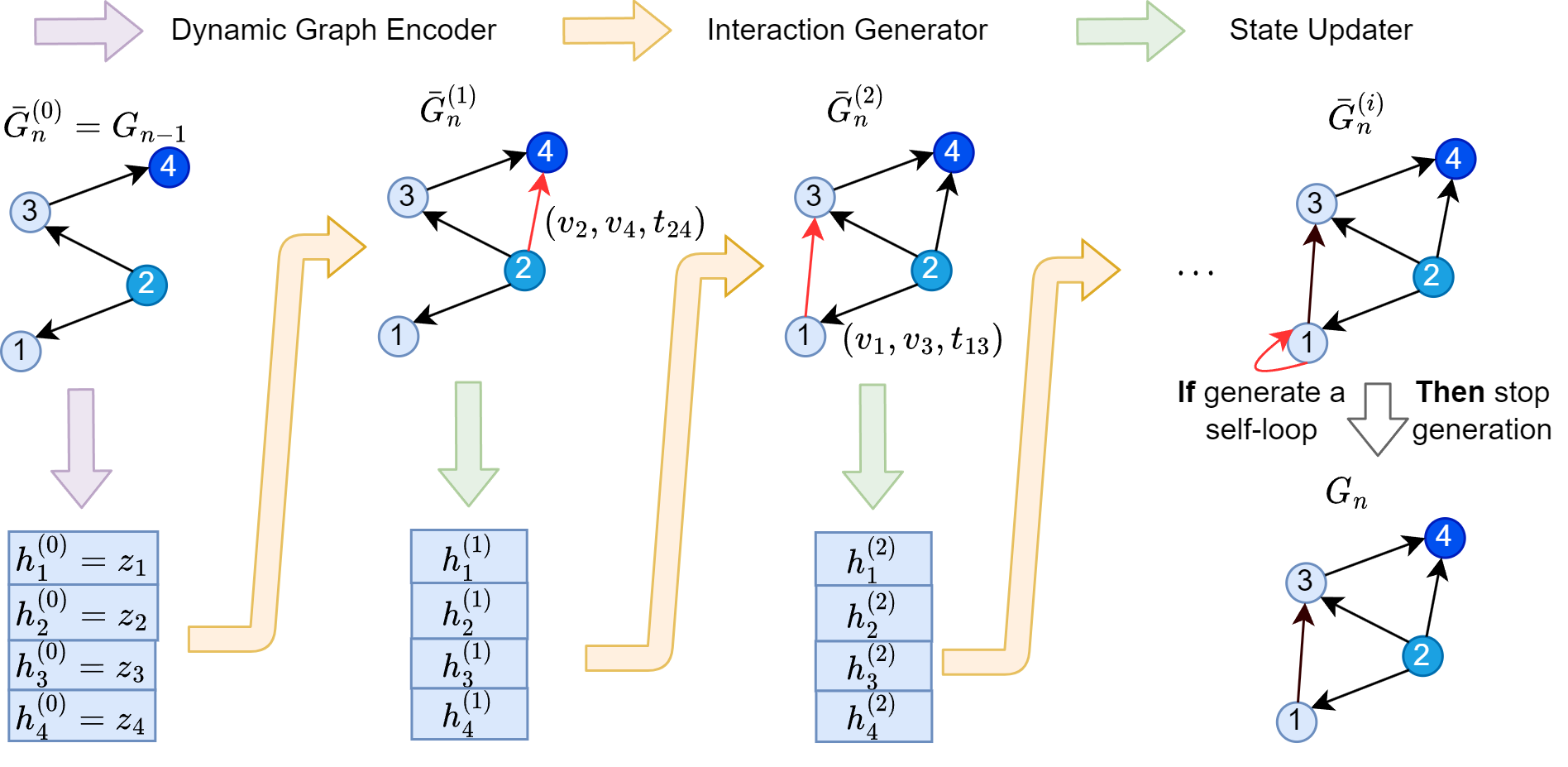}
\caption{The framework of Transportation Foundation Model (TFM).  The assumption of the TFM is very simple, that is, to regard the overall transportation system as a system with Markovian properties. The model predicts the next state $\mathcal{G}_{n}$ based on the entire system state of the previous time step $\mathcal{G}_{n-1}$. TFM is built upon graph learning approaches, with the ability to handle multi-scale and multi-source data through flexible graph structure. This flexibility enables TFM to be used for a wide range of traffic-related tasks, making it a powerful tool for accurate and realistic simulations of urban transportation systems. }
\label{fig:Framework}
\end{figure*}

\subsection{Dynamic Graph Encoder}
\label{sec:Encoder}
The function of this module is to learn the probability function $p(\mathbf{Z}_n | \mathcal{G}_n)$, which is specifically used to generate the feature embedding matrix $\mathbf{Z}_n = \{\mathbf{z}_{n,v}, \forall v \in \mathcal{V}_n\}$ composed of all nodes at time n by learning the characteristic and structural information of the graph $\mathcal{G}_n$. For the sake of convenience, we will omit all subscript labels $*_n$ in the description of this section.

Assuming we have a GNN n.etwork with $L = \{1,\dots,l,\dots,L\}$. For any node $v$, the Encoder model aims to generate its embedding $\mathbf{z}{v}$ by adopting an attention-based method to aggregate information from neighboring nodes $\mathbf{z}_{u} \in \mathcal{N}^v$, while considering the time of each interaction event and using a time encoding method $\Phi$ for representation.

\begin{figure*}[ht!]
\centering
\includegraphics[width=0.7\linewidth]{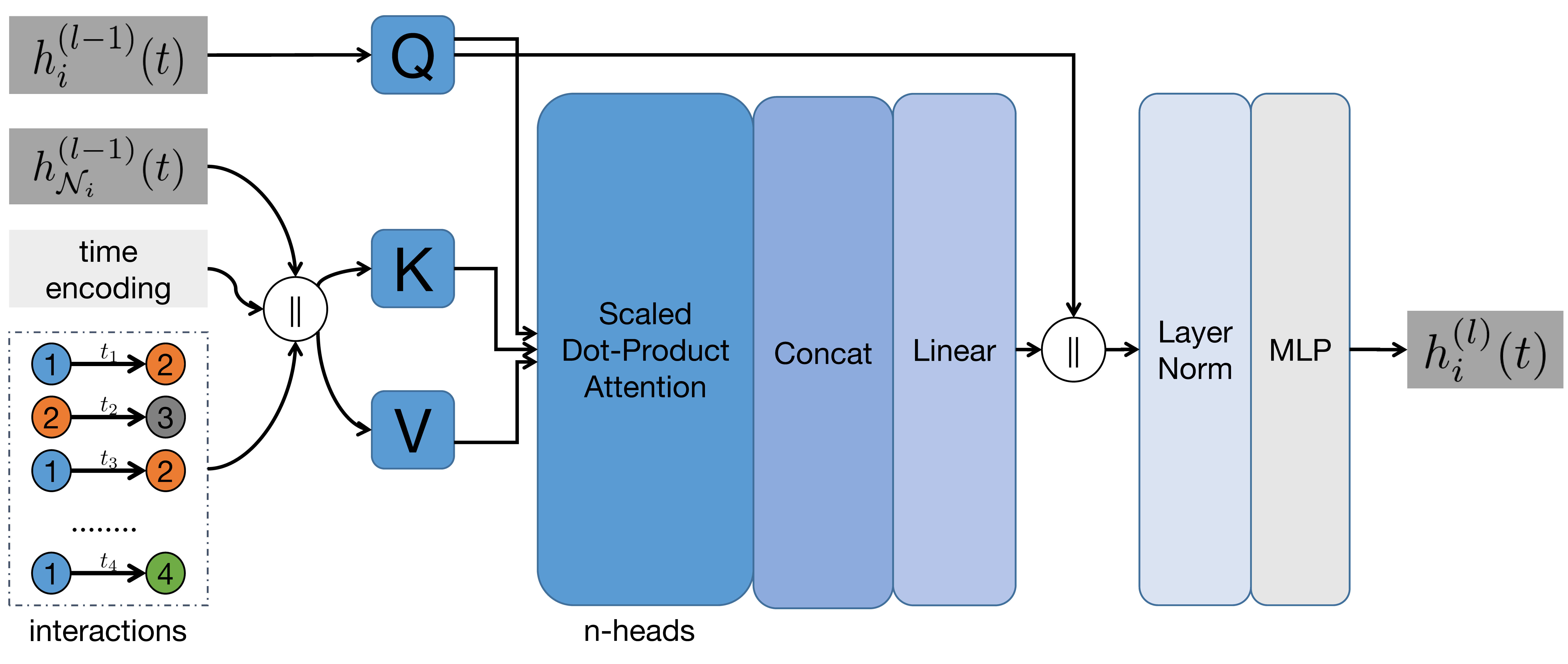}
\caption{The network structure of TFM's Graph Encoder. The Encoder Layer should be capable of encoding relational dependencies, timestamps, and optionally edge features at the same time. The encoded node representations are used by the forecaster in next section to generate the predicted Interactions.}
\label{fig:Encoder}
\end{figure*}

\begin{equation}
\label{eq:TGAT}
\mathbf{z}_v^{(l)}=f_{\text {agg }}^{\text {enc }}\left(\mathbf{z}_v^{(l-1)},\left\{g_{\text {msg }}^{\text {enc }}\left(\mathbf{z}_u^{(l-1)}, \phi\left(t_n-t\right)\right):\left(u, t\right) \in \mathcal{N}^v\right\}\right)
\end{equation}

The specific calculation form of formula~\ref{eq:TGAT} is as follows:
\begin{equation}
\begin{aligned}
\mathbf{z}_v^{(0)} & = \operatorname{MLP} (\mathbf{s}_{v}) \\ 
\mathbf{z}_v^{(l)} & =\operatorname{MLP}\left(\mathbf{z}_v^{(l-1)} \| \tilde{\mathbf{z}}_v^{(l)}\right) \\
\tilde{\mathbf{z}}_v^{(l)} & =\operatorname{Attn}^{(l)}\left(\mathbf{q}_v^{(l)}, \mathbf{K}^{(l)}, \mathbf{V}^{(l)}\right) \\
\mathbf{q}_v^{(l)} & =\left[\mathbf{z}_v^{(l-1)} \| \Phi(0)\right] \\
\mathbf{K}^{(l)} & =\mathbf{V}^{(l)} 
 =\left[\mathbf{z}_u^{(l-1)}\| \Phi\left(t_n-t\right), u \in \mathcal{N}^v\right] \\
 \Phi(\Delta{t}) &= \frac{1}{\sqrt{d_w}}\cos(\vec{\mathbf{w}}{\Delta{t}}+ \vec{\mathbf{b}})
\end{aligned}
\end{equation}

where $Attn^{(l)}$ is the short of Attention layers, which is calculated by the following equation:
\begin{equation}
\operatorname{Attn}^{(l)}\left(\mathbf{q}_v^{(l)}, \mathbf{K}^{(l)}, \mathbf{V}^{(l)}\right) =\frac{1}{|\mathcal{N}^v|} \sum_u^{\forall \mathcal{N}^v} \alpha\left(\mathbf{q}_v^{(l)}, \mathbf{K}_u^{(l)}\right) \cdot \mathbf{W}_{V}^{(l)} \mathbf{V}_{u}^{(l)}
\end{equation}

$\alpha$ represents the computational mapping function of attention weight.
\begin{equation}
\begin{aligned}
\alpha \left(\mathbf{q}_v^{(l)}, \mathbf{K}_u^{(l)}\right) & = \mathop{SoftMax}\limits_{\forall u \in \mathcal{N}^v}\left(\frac{(\textbf{W}_{Q}^{(l)}\textbf{q}_v^{(l)})(\textbf{W}_{K}^{(l)}\textbf{K}_u^{(l)})}{\sqrt{d_q}}\right) \\ 
&= \frac{ \exp\left(\frac{\left(\mathbf{W}_{Q}^{(l)} \mathbf{q}_v^{(l)}\right) \cdot \left(\mathbf{W}_{K}^{(l)} \mathbf{K}_{u}^{(l)}\right)}{\sqrt{d_q}} \right)}{\sum_j^{\forall \mathcal{N}^v} \exp\left( \frac{\left(\mathbf{W}_{Q}^{(l)} \mathbf{q}_v^{(l)}\right) \cdot \left(\mathbf{W}_{K}^{(l)} \mathbf{K}_{j}^{(l)}\right)}{\sqrt{d_q}} \right) }
\end{aligned}
\end{equation}
\subsection{Hierarchical-Probability Interaction Generator}
Due to the complexity of solving the joint probability, the conditional probability of a single edge is further split into the joint probabilities of the edge by using the hierarchical probability chain method, that is, each behavior of each traffic participant is modeled separately. It can be expressed mathematically as follows:
\begin{equation}
\begin{aligned}
p\left(u_i, v_i \mid \mathbf{Z}_n \right)=p\left(v_i \mid u_i,\mathbf{Z}_n\right) \cdot p\left(u_i \mid \mathbf{Z}_n\right) \\ 
p\left(u_i \mid \mathbf{Z}_n\right) = \mathop{SoftMax}\limits_{\forall u \in \mathcal{V}_n}\left(\mathrm{MLP}_\text{u} \left(\textbf{z}^{(i-1)}_{u} \right)\right) \\
p\left(v_i \mid u_i,\mathbf{Z}_n\right) = \mathop{SoftMax}\limits_{\forall v \in \mathcal{V}_n}\left(\mathrm{MLP}_\text{v} \left(\textbf{z}^{(i-1)}_{u_{i}} \| \textbf{z}^{(i-1)}_{v}\right)\right)
\end{aligned}
\end{equation}

Based on the probabilistic output of the neural network, we generated $(u_i, v_i)$ to represent an edge. Generally speaking, when an event generates an edge, the two end nodes corresponding to the event should change state, and then the state change will lead to the addition of new edges, which is a continuously iterative process.
\begin{figure*}[htp]
\centering
\includegraphics[width=0.8\linewidth]{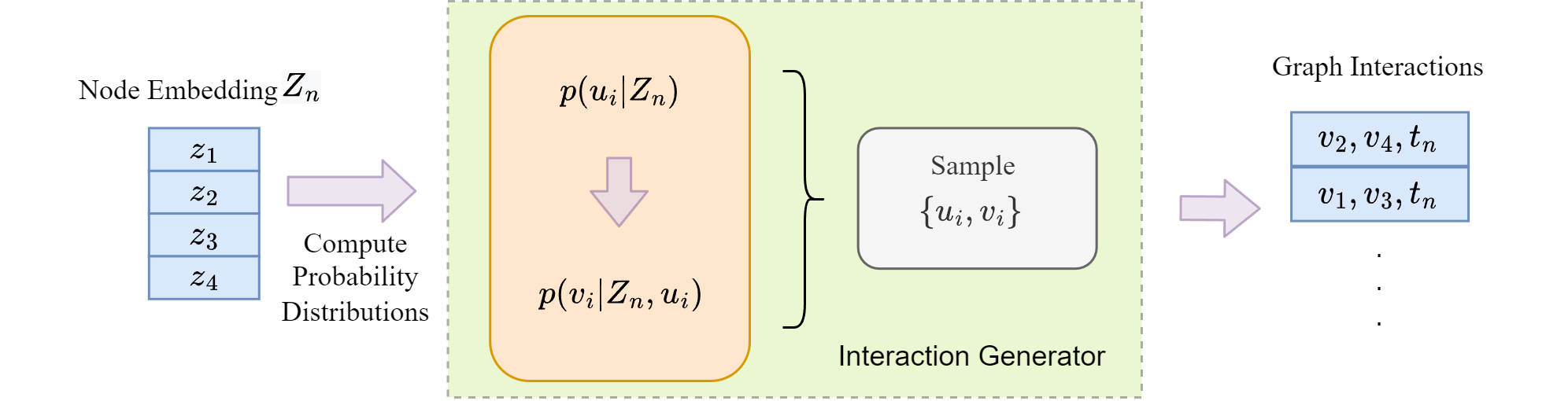}
\caption{The hierarchical probability-chain forecaster and its workflow relationship with the auto-regressive message passing module. The node embeddings are learned from the GNN Encoder described in Section \ref{sec:Encoder}.  This module is used to generate interactions between traffic particapants in the next evolution time step.}
\label{fig:Generator}
\end{figure*}

\subsection{Auto-regressive Node State Updater}
During the process of traffic simulation, the state of each traffic participant can change at any time, that is, the node state on the graph will change as time evolves. The function of this module is to learn the probability of node state changes. The change in node state may come from the influence of other nodes, or may come from the intention of the node itself, or even from the state change of the neighbors of the node. In summary, there are three main factors that influence the evolution of node state: 1. Whenever an interaction occurs, the states of the two end nodes will change. 2. Whenever an interaction is observed indirectly by other nodes, the states of the other nodes themselves may change. 3. Even if the node itself has not interacted with any other node, its own state may change over time.

Therefore, we also adopted the graph neural network method to model the impact of the first two situations on the state of the node, using a recurrent neural network, GRU, to handle the natural evolution of the node's state over time.
\begin{equation}
\begin{aligned}
\mathbf{h}_v^{(i)} & =\mathbf{s}_v^{(i-1)} \\
\mathbf{h}_v^{(i)} & =f_{\text {agg }}^{\text {upd }}\left(\left\{g_{\mathrm{msg}}^{\text {upd }}\left(\mathbf{h}_u^{(i)}, \mathbf{h}_v^{(i)}\right): u \in \mathcal{N}_{\tilde{G}^{(i)}}^v\right\}\right) \\
\mathbf{s}_v^{(i)} & =\operatorname{GRU}\left(\left[\mathbf{h}_v^{(i)} \| \phi\left(\Delta t_{n+1}\right)\right], \mathbf{z}_v^{(i-1)}\right)
\end{aligned}
\end{equation}
\subsection{Learning Objects}
Assuming that the model's prediction for node connection is ${\bar{\mathcal{E}}_{n}}$ and the prediction for node state is ${\bar{\mathbf{S}}_{n}}$, these two components together form the overall predicted graph ${\bar{\mathcal{G}}_{n}}$. Therefore, the optimization goal of the model should be to minimize the difference between the predicted graph ${\bar{\mathcal{G}}_{n}}$ and the actual graph ${\mathcal{G}_{n}}$, which can be further decomposed into two parts - the difference in graph structure and the difference in node state.
\begin{equation}
\mathcal{L} = \| {\mathcal{G}_{n}} - {\bar{\mathcal{G}}_{n}}\| =\| {\mathcal{E}_{n}} - {\bar{\mathcal{E}}_{n}}\|+ \| {\mathbf{S}_{n}} - {\bar{\mathbf{S}}_{n}}\|  
\end{equation}
\subsubsection{Structure loss}
As minimizing the difference between graphs is highly complex and not differentiable, we can naturally induce model learning metrics using a probabilistic loss function. Let ${\mathcal{G}_{n}}$ be $\mathcal{T}$ and ${\mathcal{G}_{n-1}}$ be $\mathcal{S}$. The process of generating graphs can be expressed as a sequence of events in a certain order, so the structure loss can be set as the negloglikelihood of all graph events.
\begin{equation}
\begin{aligned}
 p(\mathcal{T} \mid \mathcal{S}) 
= \prod_i & p\left(v_i \mid \mathcal{T}_{i-1}, \mathcal{S}\right) \\ \prod_{v_j \in \mathcal{T}_{i-1}} & p\left(e_{j i} \mid<v_j, v_i>, \mathcal{T}_{i-1}, \mathcal{S}\right)
\end{aligned}
\end{equation}
\begin{equation}
\mathcal{L}_{Struc} = \operatorname{NegLikelyHood}(\mathcal{T} \mid \mathcal{S})=-\log p(\mathcal{T} \mid \mathcal{S}) 
\end{equation}
\subsubsection{State loss}
As the node states are dense feature vectors, comparing them is very easy. We can simply use the mean square error (MSE) loss function to calculate the differences between the node states on the graph ${\bar{\mathcal{G}}_{n}}$ and the true graph ${\mathcal{G}_{n}}$, and obtain gradients.
\begin{equation}
\mathcal{L}_{State} = \operatorname{MSE}({\mathbf{S}_{n}} , {\bar{\mathbf{S}}_{n}}) = \sum_{v \in \{\bar{\mathcal{V}}_{n} \cap {\mathcal{V}}_{n}\}} \| {\mathbf{s}_{v}} - {\bar{\mathbf{s}}_{v}}\|^{2}
\end{equation}

\section{Case Study: A city-scale simulation}

\subsection{Data Collection}
\begin{figure*}[t]
\centering
\includegraphics[width=0.7\linewidth]{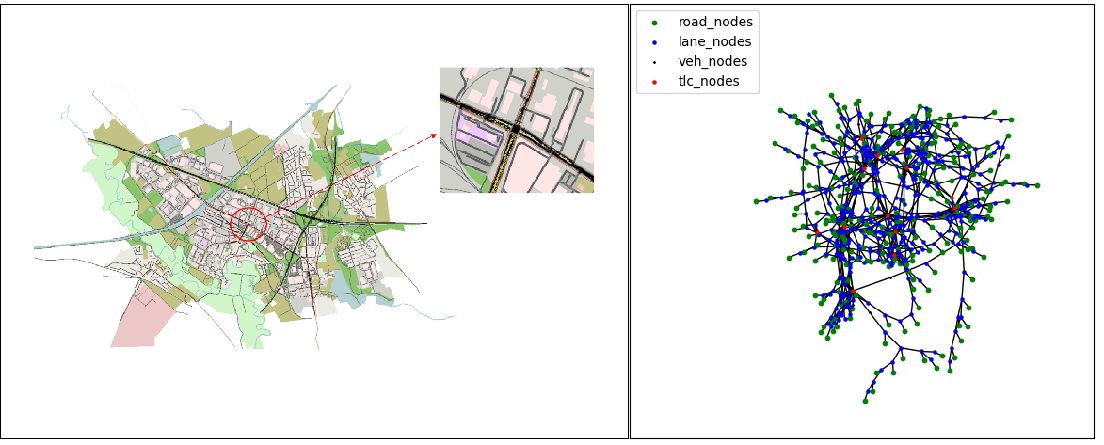}
\caption{ Left: The city of Bologna's street network. Right: The graph structure that represents the traffic system in the city of Bologna and illustrates the links between different system agents. The graph comprises nodes that represent roads, lanes, vehicles, and traffic lights within the city, and edges that depict relationships between these nodes. This graph structure facilitates a more comprehensive understanding of the transportation network by offering a visual representation of the system.}
\label{fig_graph_of_city}
\end{figure*}

Effective transportation planning and decision-making requires accurate and reliable simulations of large-scale traffic scenarios. Such simulations serve to provide policymakers and engineers with a platform to evaluate traffic management strategies and infrastructure improvements. In this study, we propose the TFM, and its performance is demonstrated through a simulation of Bologna city, which can be seen in Figure \ref{fig_graph_of_city}.

Bologna is a vast metropolis, similar in complexity to the SUMO example \cite{behrisch_traffic_2015}. The simulation accurately models the city's roads, intersections, traffic lights, and a vast number of vehicles, creating a simulation that represents real-world traffic conditions. The simulation scenario utilized consists of 11,079 vehicles, 26 traffic controllers, 268 roads, and 412 lanes, enabling the generation of a wealth of data that can be used to evaluate the effectiveness of various traffic management strategies.

\subsection{Results}
The MFDs generated by TFM and SUMO both reveal consistent trends with respect to the relationship between traffic density and speed. However, while SUMO's MFD curves present a consistent and tight pattern, TFM's MFD curves expose a higher variance of data points. This disparity may be attributed to TFM's data-driven methodology, which allows for more precise and realistic modeling of complex traffic behaviors across diverse regions within a city. In contrast, SUMO assumes traffic behavior is uniform across all roads, a method which may not always be applicable in real-world traffic. The study~\cite{laval_stochastic_2015} utilized SUMO outputs to optimize the TFM model, which may cause limitations in modeling the many nuanced complexities of real-world traffic. As such, further scientific investigation is essential, utilizing real-world data to establish how the simulation can be refined to reflect local traffic behavior's intricacies and variability.

\begin{figure}[ht]
\centering
\includegraphics[width=1\linewidth]{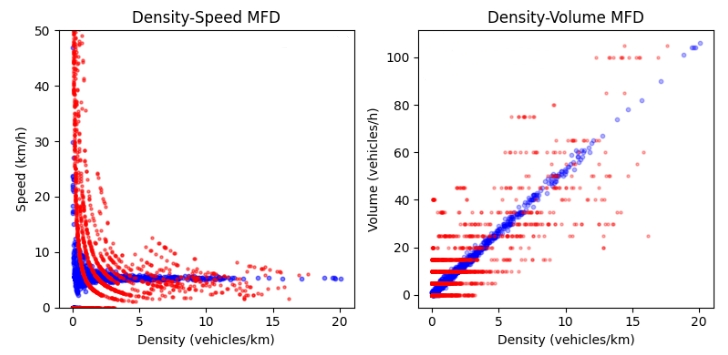}
\caption{Comparison of the MFD between SUMO and TFM. The blue color represents the results of the SUMO simulation software, while the red dots represent the results of the proposed TFM. TFM method shows more diversity in traffic patterns, because the TFM method can recover diverse traffic patterns from a large amount of data of different scenarios, and even new patterns can emerge, instead of solely relying on simple physical rules like SUMO does.}
\label{fig_s2-mfd}
\end{figure}

\begin{figure}[ht]
\centering
\includegraphics[width=1\linewidth]{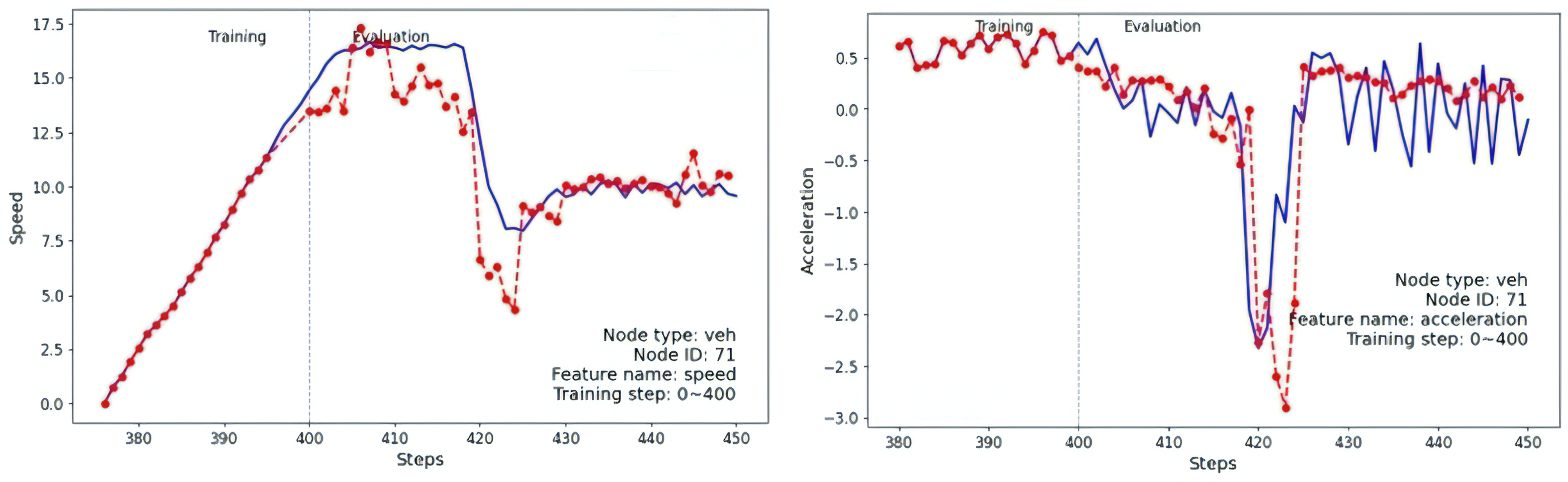}
\caption{Evaluating the TFM method from the micro perspective of traffic vehicles. The blue curve represents the results of SUMO, while the red curve represents the results of TFM. Although the relationship between speed and acceleration is not one-to-one in the prediction results of TFM, we can see that TFM has learned the control behavior pattern of vehicles. In practical applications, people can focus on one of the indicators predicted by TFM and derive other indicators through physical rules.}
\label{fig_pred}
\end{figure}

In Figure \ref{fig_pred}, we will examine the understanding and prediction of micro traffic patterns in the TFM model from the perspective of transportation participants. We can see that TFM accurately learns the behavioral patterns to be produced by individual traffic participants. When the model predicts many traffic participants simulataneously, interesting global patterns can emerge because the graphic model can model the interaction among them. For example, the scattered MFD shown in Figure \ref{fig_s2-mfd}.

\section{CONCLUSIONS}

In conclusion, the Transportation Foundation Model (TFM) proposed in this paper is a novel approach that integrates traffic simulation principles into traffic prediction. This model addresses the challenges faced by traditional systems and provides a promising tool for solving complex transportation problems with real data. By using graph structures and dynamic graph generation algorithms, TFM captures the participatory behavior and interaction of transportation system actors to produce accurate predictions of traffic outcomes in densely populated urban areas. As a data-driven and model-free simulation method, TFM has the ability to understand and efficiently utilize big data, making it a promising option for optimizing the efficiency of traffic management in modern cities.




\section*{Data Availability}
The SUMO simulation platform and data is publicly available at
\url{https://www.eclipse.org/sumo}.
.

\section*{Code availability}
The code of TransWorldNG was implemented in Python using the deep learning framework of PyTorch. Code, trained models, and scripts reproducing the experiments of this paper are available at \url{https://github.com/PJSAC/TransWorldNG}.

\bibliographystyle{IEEEtran}
\bibliography{IEEEabrv}

\end{document}